\documentclass[runningheads]{llncs}

\usepackage{eccv}

\usepackage{eccvabbrv}

\usepackage{graphicx}
\usepackage{booktabs}

\usepackage[accsupp]{axessibility}  

\usepackage{hyperref}

\usepackage{orcidlink}

\usepackage{multirow}
\usepackage[ruled,lined]{algorithm2e}
\usepackage{pifont}
\newcommand{\cmark}{\ding{51}}%
\newcommand{\xmark}{\ding{55}}%
\usepackage{float}
\usepackage{colortbl}

\begin{document}

\title{Simplifying Source-Free Domain Adaptation for Object Detection: Effective Self-Training Strategies and Performance Insights} 

\titlerunning{Simplifying Source-Free Domain Adaptation for Object Detection}

\author{Yan Hao\inst{1} \and
Florent Forest\inst{1}\orcidlink{0000-0001-6878-8752} \and
Olga Fink\inst{1}\orcidlink{0000-0002-9546-1488}}

\authorrunning{Y.~Hao et al.}

\institute{Intelligent Maintenance and Operations Systems, EPFL, 1015 Lausanne, Switzerland\\
\email{first.last@epfl.ch}\\
\url{https://imos.epfl.ch}}

\maketitle

\begin{abstract}
    This paper focuses on source-free domain adaptation for object detection in computer vision. This task is challenging and of great practical interest, due to the cost of obtaining annotated data sets for every new domain. 
    Recent research has proposed various solutions for Source-Free Object Detection (SFOD), most being variations of teacher-student architectures with diverse feature alignment, regularization and pseudo-label selection strategies.
    Our work investigates simpler approaches and their performance compared to more complex SFOD methods in several adaptation scenarios. We highlight the importance of batch normalization layers in the detector backbone, and show that adapting only the batch statistics is a strong baseline for SFOD. We propose a simple extension of a Mean Teacher with strong-weak augmentation in the source-free setting, Source-Free Unbiased Teacher (SF-UT), and show that it actually outperforms most of the previous SFOD methods. Additionally, we showcase that an even simpler strategy consisting in training on a fixed set of pseudo-labels can achieve similar performance to the more complex teacher-student mutual learning, while being computationally efficient and mitigating the major issue of teacher-student collapse.
    We conduct experiments on several adaptation tasks using benchmark driving datasets including (Foggy)Cityscapes, Sim10k and KITTI, and achieve a notable improvement of 4.7\% AP50 on Cityscapes$\rightarrow$Foggy-Cityscapes compared with the latest state-of-the-art in SFOD. Source code is available at \href{https://github.com/EPFL-IMOS/simple-SFOD}{https://github.com/EPFL-IMOS/simple-SFOD}.
  \keywords{Source-free domain adaptation \and Object detection}
\end{abstract}

\section{Introduction}

Domain adaptation aims to transfer knowledge acquired from a source domain to a related but differently distributed target domain, characterized by domain shift. Source-free domain adaptation (SFDA) addresses a more challenging scenario where only a pre-trained model from the source domain and unlabeled data from the target domain are accessible. This stands in contrast to the standard unsupervised domain adaptation (UDA), where labeled source data is available. SFDA becomes particularly valuable in practical situations where obtaining labels for target-domain data is difficult, and source data usage is restricted due to privacy concerns, storage limitations, or deployment constraints. We focus on Source-Free Object Detection (SFOD), which aims to adapt a detector trained on a source domain to an unlabeled target domain without accessing the source data.

\begin{figure}[t]
    \centering
    \includegraphics[width=0.6\linewidth]{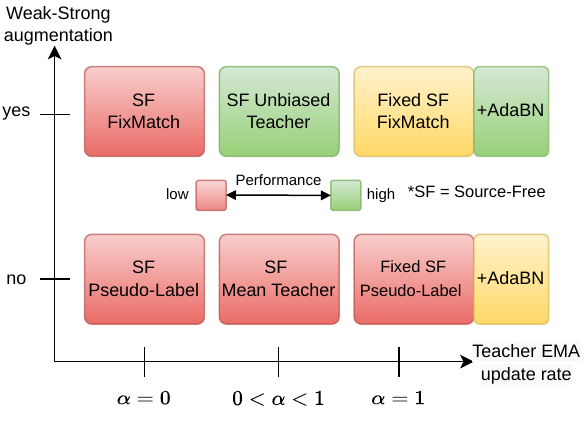}
    \caption{Overview of Source-Free Mean Teacher configurations for SFOD with different teacher update rates $\alpha$ and use of weak-strong augmentation. The extreme case of $\alpha=0$ (i.e., teacher = student) corresponds to (source-free) Pseudo-Label \cite{lee_pseudo-label_2013} and FixMatch \cite{sohn2020fixmatch} respectively. On the other hand, $\alpha=1$ boils down to freezing the teacher and training on a fixed set of pseudo-labels. Surprisingly, training on fixed pseudo-labels after AdaBN \cite{li2018adaptive} yields similar performance than more complex teacher-student mutual learning and challenges state-of-the-art SFOD methods.}
    \label{fig:overview}
\end{figure}


Object detection, the task of simultaneously localizing and classifying multiple objects in an image, is a major field in computer vision with numerous real-world applications. While deep learning-based object detectors have demonstrated remarkable success in recent years \cite{ren2015faster}, their performance often experiences significant degradation in the presence of domain shifts. Domain adaptation for object detection is notably more challenging than for classification, as it necessitates not only accurate classification but also precise localization. In addressing this challenge, various approaches have been proposed to tackle UDA for object detection tasks (UDAOD) \cite{su2020adapting,he2020domain,li2020spatial,zhao2020collaborative,vs2021mega,zhang2021self,deng2021unbiased,zhao_task-specific_2022,li2022cross}.

Despite its practical significance, SFOD has received comparatively less attention. The task becomes particularly demanding when there is a significant domain shift, and the source data is inaccessible,  preventing the explicit reduction of the domain shift. A majority of the proposed SFOD methods adopt the Mean Teacher (MT) framework with self-training on confident pseudo-labels \cite{li2021free,huang2021model,xiong2021source,li2022source,wei_entropy-minimization_2023,chu_adversarial_2023,vs_instance_2023,zhang2023refined,chen_exploiting_2023,liu_periodically_2023}. Originally introduced for semi-supervised learning (SSL), Mean Teacher \cite{tarvainen2017mean} is a variant of temporal ensembling \cite{laine_temporal_2017} where knowledge is distilled from a teacher network into a student network. The student receives pseudo-labels from the teacher and is updated with standard gradient-based learning, while the teacher is gradually updated through an exponential moving average (EMA) of the previous student weights. This approach aims to increase the robustness to inaccurate and noisy predictions on the unlabeled target data.

A major issue in fully-unlabeled training of teacher-student architectures is the collapse of teacher and student during training. Concretely, whenever the performance of the teacher starts to degrade in the target domain, it causes an amplified degradation of the student, which is turns leads to a performance collapse in both networks. To alleviate this challenge, \cite{chu_adversarial_2023} increases the period of teacher updates to 2500 steps (without specifying the update rate, or how the update period was set). PETS \cite{liu_periodically_2023} proposes a periodic exchange of teacher and student and an additional teacher with slower updates to effectively prevent catastrophic collapse. However, this increases the complexity of the overall architecture and necessitates to set an exchange period empirically.

In this work, we demonstrate the effectiveness of simpler approaches in SFOD compared to the previously mentioned more complex SFOD methods. Firstly, we emphasize the importance of batch normalization (BN) \cite{ioffe_batch_2015} layers, showcasing their impact on adaptation using a more modern backbone, VGG16-BN (with BN layers), instead of the VGG16 used in previous lines of work. Specifically, adapting only the batch statistics on unlabeled target training data, a technique known as AdaBN \cite{li2018adaptive}, proves to be a strong baseline for SFOD. Secondly, we propose a straightforward extension of the Unbiased Teacher (UT) \cite{liu2021unbiased} to the source-free setting, called Source-Free Unbiased Teacher (SF-UT). Finally, we explore various configurations of self-training strategies for SFOD (see Figure~\ref{fig:overview}). This includes different teacher EMA update rates, weak-strong augmentation usage, and batch statistics adaptation. We investigate extreme cases of the teacher update rate $\alpha$, ranging from $\alpha=0$ (equivalent to a source-free version of Pseudo-Label training (SF-PL) \cite{lee_pseudo-label_2013} using solely unlabeled data)  to $\alpha=1$ (corresponding to a fixed teacher, i.e. training on the fixed set of initial pseudo-labels produced by the source-trained model). When SF-PL is combined with weak-strong augmentation, it becomes equivalent to a source-free version of FixMatch \cite{sohn2020fixmatch} (SF-FM). For the latter cases with $\alpha=1$ (i.e., fixed pseudo-labels), we refer to them as Fixed SF-PL and Fixed SF-FM, for cases with and without weak-strong augmentation, respectively. We show that the approach AdaBN + Fixed SF-FM, using fixed initial pseudo-labels generated by a source model adapted to the target domain with AdaBN beforehand, is almost as effective as the fully-fledged Unbiased Teacher \cite{liu2021unbiased} adapted for the source-free setting (our proposed SF-UT). Our proposed method AdaBN + Fixed SF-FM offers the advantage of stable training, as it has no feedback loop between the teacher and the student, while achieving similarly good performance.

The contributions of this work can be summarized as follows:
\begin{itemize}
    \item We emphasize the significance of batch normalization layers and demonstrate the effectiveness of batch statistics adaptation for source-free object detection tasks.
    \item We introduce a source-free extension of the Unbiased Teacher (SF-UT).
    \item We propose a novel, lightweight strategy that combines AdaBN with training on a set of fixed pseudo-labels using weak-strong augmentations (AdaBN + Fixed SF-FixMatch).
    \item We conduct experiments in three SFOD adaptation scenarios: adverse weather adaptation (Cityscapes$\rightarrow$Foggy-Cityscapes), cross-camera adaptation (KITTI $\rightarrow$ Cityscapes) and synthetic-to-real adaptation (SIM10k$\rightarrow$Cityscapes). We demonstrate the superior or closely comparable performance of SF-UT compared to more complex state-of-the-art SFOD methods. In addition, the simple AdaBN + Fixed SF-FixMatch strategy achieves competitive results and avoids the common collapse in teacher-student methods.
\end{itemize}



\section{Related work}

%
\subsection{UDA for Object Detection (UDAOD)}

The field of domain adaptation aims to transfer learned knowledge from a source domain to a target domain. In Unsupervised Domain Adaptation (UDA), labeled data is available in the source domain and only unlabeled samples are available in the target domain. The main principle shared by most UDA methods is to explicitly reduce the domain shift while jointly performing supervised training on the source data. The reduction of domain discrepancies can be achieved by matching moments of the source and target feature distributions \cite{long2015learning}, optimal transport \cite{courty_joint_2017} or learning domain-invariant features through kernel learning \cite{duan2012domain,gong2012geodesic} or domain-adversarial training \cite{ganin_domain-adversarial_2016,long_conditional_2018,rangwani_closer_2022}.


In UDA for Object Detection tasks (UDAOD), one prevalent approach involves aligning the features between the source and target domains at the image or the instance level via domain-adversarial training, including DA-Faster \cite{chen2018domain}, SW-Faster \cite{saito_strong-weak_2019}, SSA-DA \cite{zhao2020bi}, ICR-CCR \cite{xu2020exploring}, SGA-S \cite{zhang2021self}, ATF \cite{he2020domain}, MeGA-CDA \cite{vs2021mega} and CST-DA \cite{zhao2020collaborative}. Methods employing pseudo-labels on the target domain have also been investigated. For instance, NL \cite{khodabandeh2019robust} designed a robust-to-noise training scheme for object detection which is trained on noisy bounding boxes and labels acquired from the target domain as pseudo-ground-truth. A Mean Teacher (MT) architecture is adopted in MTOR \cite{cai2019exploring}, UMT \cite{deng2021unbiased} and AT \cite{li2022cross}. An attention mechanism to focus on the most discriminative features is leveraged in SAPNet \cite{li2020spatial}.

\subsection{Source-Free Object Detection (SFOD)}
%

Source-Free Object Detection (SFOD) is challenging due to the impossibility of explicitly reducing the domain discrepancy in absence of source data. Recently, several methods have been proposed to address this problem. As only unlabeled target data is available, they employ self-training strategies with pseudo-labeling of target samples. SED \cite{li2021free} employs a self-entropy descent policy to obtain the suitable confidence threshold for pseudo-labeling. HCL \cite{huang2021model} explores memory-based learning and proposes a historical contrastive learning method for instance discrimination and category discrimination. Recent methods are based on the Mean Teacher paradigm in combination with various additional alignment, pseudo-label selection or regularization strategies. SOAP \cite{xiong2021source} uses adversarial learning to transfer the detector by perturbing target images with domain-specific noise. In a similar spirit, LODS \cite{li2022source} learns the domain shifts by enhancing the style of each target domain image and leveraging the style degree difference between the original and the enhanced image to guide adaptation. A$^2$SFOD \cite{chu_adversarial_2023} proposes an approach in four stages and divides the target data based on a variance criterion and aligns their features via adversarial training. IRG \cite{vs_instance_2023} adds a graph-guided constrastive loss based on learning instance relations with a graph convolution network. ESOD \cite{wei_entropy-minimization_2023} uses entropy minimization to find an optimal confidence threshols. In RPL \cite{zhang2023refined}, the MT architecture is harnessed with a category-aware adaptive threshold for pseudo-labeling and a localization-aware pseudo-label assignment strategy. Chen \textit{et al.} \cite{chen_exploiting_2023} introduces a second confidence threshold for low-confidence proposals used in a spatial contrastive loss term. Finally, PETS \cite{liu_periodically_2023} addresses the issue of collapse in teacher-student mutual learning by periodically exchanging teacher and student, as well as adding an additional dynamic teacher with slower updates to stabilize the training.

\subsection{Normalization layers adaptation}

Normalization layers such as Batch normalization (BN) \cite{ioffe_batch_2015} or Instance normalization (IN) \cite{ulyanov_instance_2017} are widely used to train deep neural networks. BN consists in normalization and scaling the features in each batch using affine parameters $\gamma$ and $\beta$, i.e.:
\begin{equation}
    \text{BN}(\mathbf{x}; \gamma, \beta) = \gamma \cdot \frac{\mathbf{x} - \mu}{\sqrt{\sigma^2 + \epsilon}} + \beta,
\end{equation}
where $\mu$ and $\sigma^2$ are the mean and variance statistics of feature activations in the batch $\mathbf{x}$. During training, BN maintains running estimates of the batch statistics, updated through EMA. Traditionally, statistics from the training set are used at inference time, which is an issue in case of domain shift. Several approaches have proposed to modulate normalization layers across the network using the target domain, at the advantage of requiring no or few additional parameters. \cite{pan_two_2018} has shown how BN could preserve content information while IN learns features invariant to appearance changes. Domain-specific BatchNorm \cite{chang_domain-specific_2019} proposes using separate BN parameters for each domain. The best combination of BN and IN is learned in \cite{seo_learning_2020}. However, these approaches require source data. AdaBN \cite{li2018adaptive} is a UDA method for classification where batch statistics from the source domain are replaced by statistics of the target domain, estimated over batches of or the entire target test set. Target-specific normalization is also used in combination with adversarial training in \cite{zhang_generalizable_2022}. Tent \cite{wang_tent_2021} is a test-time adaptation method that additionally optimizes the affine transformations using entropy minimization. Because these approaches induce inter-image dependency at test-time, UBNA \cite{klingner_unsupervised_2021} propose to use a separate adaptation set, and progressively mix source and target-domain statistics. When the adaptation set, that we call target train set, is large enough, it becomes equivalent to AdaBN over the entire target train set, which is the method we adopt in this work.
Surprisingly, previous SFOD lines of work have been using backbones such as VGG16 without BN, and have not considered normalization layer adaptation.


%
\subsection{Self-training}

Self-training on confident pseudo-labels (PLs) is an effective technique in semi-supervised learning (SSL). In Pseudo-Label \cite{lee_pseudo-label_2013}, confident predictions on the unlabeled data are used as pseudo-labels and added to the labeled data for the next training round. FixMatch \cite{sohn2020fixmatch} proposes to leverage weak-strong augmentation and use the PLs from weakly-augmented inputs as supervision targets for strongly-augmented inputs. Mean Teacher (MT) \cite{tarvainen2017mean} is a variant of temporal ensembling \cite{laine_temporal_2017} where knowledge is distilled from a teacher network into a student network via a consistency loss or pseudo-labeling. The student is updated with standard gradient-based learning, while the teacher is gradually updated through an exponential moving average (EMA) of the previous student weights, resulting in an ensemble of previous iterations of the students. This approach aims to increase the robustness to inaccurate and noisy predictions on the unlabeled target data. Unlike previous works that maintained an EMA of predictions, the EMA of weights allows for a shorter update period, as predictions change only once per epoch while weights change every step. Unbiased Teacher (UT) \cite{liu2021unbiased} proposes a Mean Teacher with weak-strong augmentation for semi-supervised object detection. After a source-only training phase, the teacher, fed with weakly augmented target inputs, generates pseudo-labels to train the student, fed with strongly-augmented inputs. Adaptive Teacher \cite{li2022cross} extended UT to unsupervised domain adaptation. Additionally, it employs adversarial learning by incorporating a discriminator to align image-level features across both domains in the student network.

\section{Method}\label{sec:method}

\begin{figure}[t]
    \centering
    \includegraphics[width=0.9\linewidth]{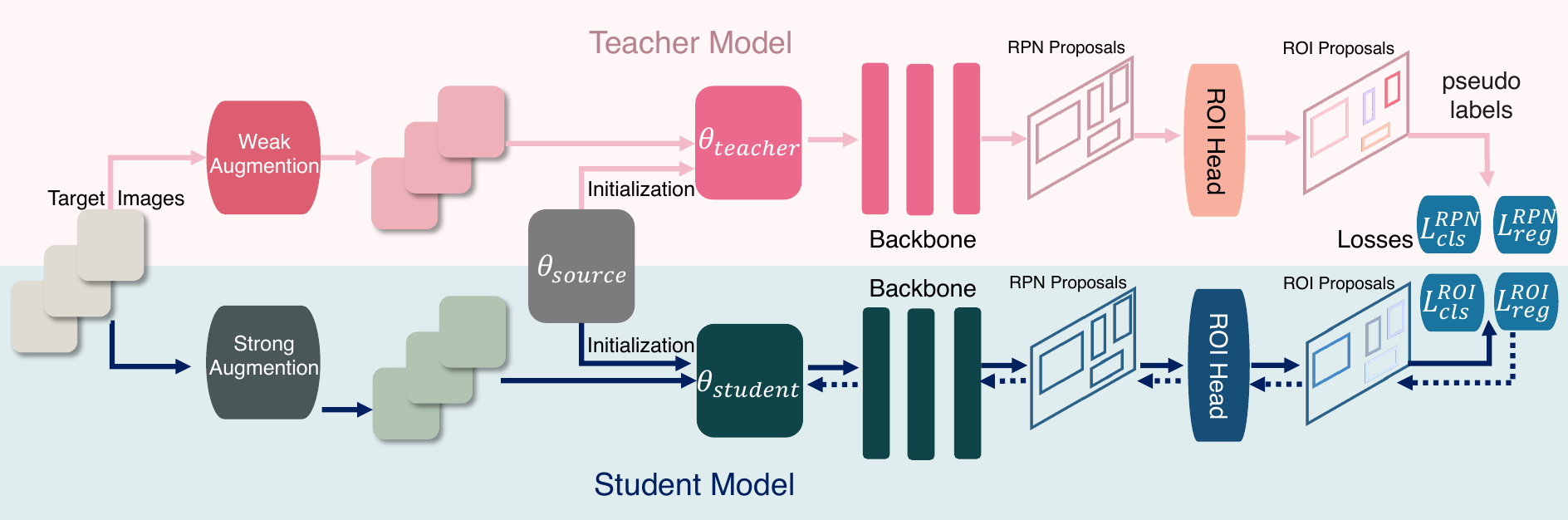}
    \caption{Proposed Source-Free Unbiased Teacher (SF-UT) architecture.
    }
    \label{fig:sfut}
\end{figure}

%
\subsection{Preliminaries}

Given a pre-trained object detection model on the source domain with parameters $\theta_{\text{source}}$, the goal is to adapt this model using only unlabeled training data $\mathbf{X}_{\text{train}} = \{\mathbf{x}_i\}_{1 \leq i \leq N}$ in the target domain in order to improve the performance on the target test data $\mathbf{X}_{\text{test}}$.



Following previous works in SFOD, the base detector is a Faster-RCNN \cite{ren2015faster}, which consists in a feature extraction backbone, a region proposal network (RPN) and a region of interest (ROI) head. For each input image $\mathbf{x}i$, the detector outputs a set of object proposals $y_i = (b_i, c_i)$ consisting of bounding boxes and categories.

\subsection{Source-Free Unbiased Teacher (SF-UT)}

We first introduce an adaptation of the semi-supervised method Unbiased Teacher (UT) \cite{liu2021unbiased} for SFOD, called Source-Free Unbiased Teacher (SF-UT) in the rest of the paper. The architecture is illustrated in Figure~\ref{fig:sfut}. The main difference with UT is that only unlabeled target data is being used, in contrast to UT where labeled source data is also used to train the student network. Let $\theta_{\text{student}}$ and $\theta_{\text{teacher}}$ denote the weights of the student and teacher networks, respectively. Both networks are initialized with the pre-trained source model, i.e.: 

\begin{equation}
    \theta_{\text{student}} \gets \theta_{\text{source}}, \theta_{\text{teacher}} \gets \theta_{\text{source}}.
\end{equation}

\subsubsection{Student training}

\begin{figure}[t]
    \centering
    \includegraphics[width=0.9\linewidth]{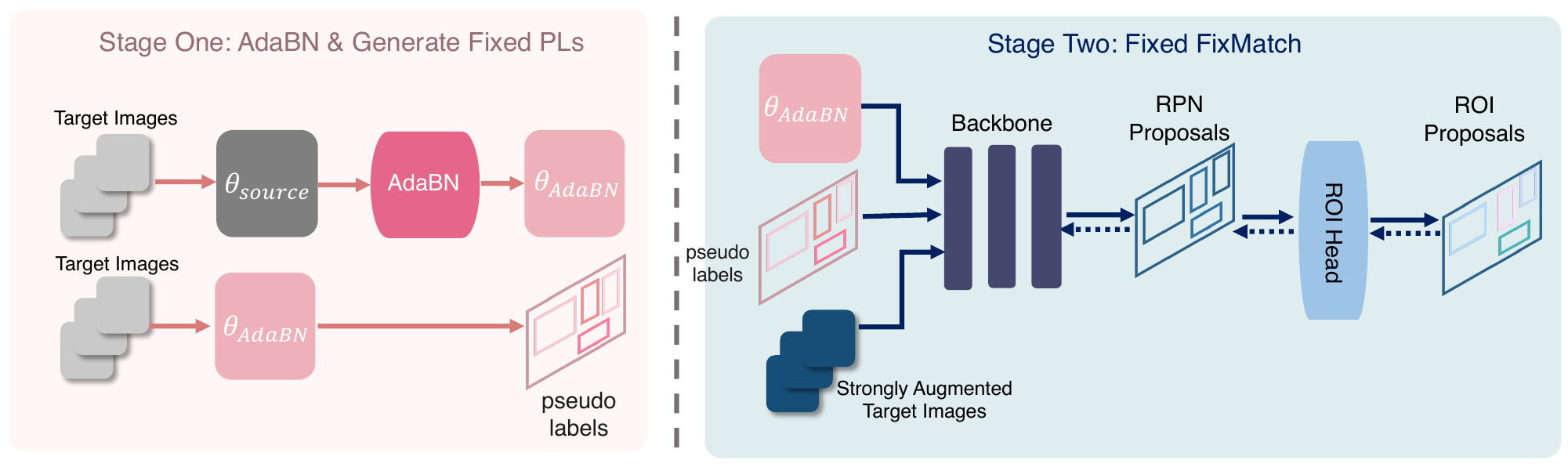}
    \caption{Proposed Fixed Source-Free FixMatch (Fixed SF-FM) strategy.}
    \label{fig:sfpl}
\end{figure}

Given a batch of unlabeled target inputs $\mathbf{x}_i$, the teacher network is used to predict a set of pseudo-labels $\hat{y}_i = (\hat{b}_i, \hat{c}_i)$ based on the weakly-augmented inputs. To avoid error accumulation due to noisy pseudo-labels, we set a fixed confidence threshold $\tau$ to filter PLs, using the confidence of the categories $\hat{c}_i$ as a proxy for correctness. Then, the student is trained on the strongly-augmented inputs $\mathbf{x}_i^s$ using the standard detection loss of Faster-RCNN, expressing as

\begin{multline}
    \mathcal{L}_{\text{SF-UT}}(\mathbf{x}_i^s, \hat{y}_i) = \mathcal{L}_{\text{cls}}^{\text{RPN}}(\mathbf{x}_i^s, \hat{y}_i) + \mathcal{L}_{\text{reg}}^{\text{RPN}}(\mathbf{x}_i^s, \hat{y}_i) \\+ \mathcal{L}_{\text{cls}}^{\text{ROI}}(\mathbf{x}_i^s, \hat{y}_i) + \mathcal{L}_{\text{reg}}^{\text{ROI}}(\mathbf{x}_i^s, \hat{y}_i)
    \label{eq:loss},
\end{multline}

where $\mathcal{L}_{\text{cls}}^{\text{RPN}}$ and $\mathcal{L}_{\text{reg}}^{\text{RPN}}$ are respectively the RPN classification and regression loss for candidate proposals, and  $\mathcal{L}_{\text{cls}}^{\text{ROI}}$ and $\mathcal{L}_{\text{reg}}^{\text{ROI}}$ are respectively the classification and regression losses for the ROI head. We use cross-entropy for the classification losses and the smooth $\ell_1$ loss for regression losses.

It is important to note that in previous works, the bounding box regression losses are not applied on the unlabeled data, due to the absence of confidence estimates for the bounding box regression \cite{liu2021unbiased,li2022cross,li2021free}. In our SF-UT, we train on the complete loss including the regression terms and show it has a positive impact in Section~\ref{sec:experiments}.

\subsubsection{Teacher updating}

Between each training iteration, the teacher's weights are updated by an EMA of the teacher's weighs with the current student's weights with an EMA update rate of $\alpha$:

\begin{equation}
    \theta_{\text{teacher}} \gets \alpha \cdot \theta_{\text{teacher}} + (1 - \alpha) \cdot \theta_{\text{student}}.
\end{equation}





%
\subsection{Source-Free Pseudo-Label \& FixMatch}

As seen previously, semi-supervised methods can be applied to the source-free setting by using fully unlabeled data. Hence, we also investigate the common SSL methods Pseudo-Label \cite{lee_pseudo-label_2013} and FixMatch \cite{sohn2020fixmatch}. We denote the source-free variants as SF-PL and SF-FM respectively. These methods can be seen as a special case of Mean Teacher with $\alpha=0$ (
i.e., teacher and student are a single model), without and with weak-strong augmentation.

In SF-PL, after initialization with the source weights, the network alternates between generating confident pseudo-labels on weakly-augmented inputs and training the detector using the loss \ref{eq:loss}. SF-FM follows an identical procedure, except that the training is performed on strongly-augmented inputs.

\subsection{Fixed SF-PL \& SF-FixMatch}

The two previously presented strategies are simple but lack the robustness of Mean Teacher, leading to a quick collapse due to noisy pseudo-labels. To alleviate this issue, we propose to remove the pseudo-labeling feedback loop by fixing the initial set of PLs and only training on these fixed PLs. This can be seen as the other extreme degenerate case of Mean Teacher with $\alpha=1$, i.e. the teacher is never updated, always producing the same PLs. As previously, we call these methods Fixed SF-PL and Fixed SF-FixMatch whenever the network is trained on weak or strongly-augmented inputs.

In order to improve the accuracy of the fixed set of PLs, we propose to initially adapt the batch statistics of the source model using AdaBN. This only requires one forward pass over several batches of target data set in practice. In our study, we make one pass over the entire target training set. This strategy is represented in Figure~\ref{fig:sfpl}.

\section{Experiments}\label{sec:experiments}


%
\subsection{Datasets}

The experiments are conducted on benchmark driving datasets, including Cityscapes \cite{Cordts2016Cityscapes}, Foggy-Cityscapes\cite{sakaridis_semantic_2018}, KITTI~\cite{Geiger2012CVPR} and SIM10k~\cite{johnson2017driving}. Following previous literature \cite{chen2018domain}, we evaluate the methods under three different domain shift scenarios:

\begin{itemize}
    \item Adverse weather adaptation from normal weather to foggy weather using Cityscapes$\rightarrow$Foggy-Cityscapes.
    \item Synthetic-to-real adaptation using SIM10k$\rightarrow$Cityscapes.
    \item Cross-camera adaptation using KITTI$\rightarrow$Cityscapes.
\end{itemize}

For Foggy-Cityscapes, we only consider the highest level of fog (0.02). For SIM10k and KITTI, we only consider the Car class in the evaluation, as usually done for this transfer task.

\subsection{Experimental settings \& Model selection}

\textbf{Detector.} We use Faster-RCNN \cite{girshick2015fast} as the base detector. The feature extraction backbone is a VGG16 \cite{simonyan2014very} initialized with ImageNet weights. It is worth noting that we are using the architecture comprising batch normalization layers (VGG16-BN), unless specified otherwise.

\noindent\textbf{Augmentations.} For weak-strong augmentation, we adopt the transformations used in \cite{liu2021unbiased,li2022cross} for fair comparison. Weak augmentation involves random horizontal flipping. Strong augmentation involves a sequence of transformations including random color jittering, grayscaling, Gaussian blurring, and cutting out patches, which is applied on top of the weak augmentation. These augmentations aim at increasing the student's robustness and generalization with respect to appearance variations and occluded objects.






\noindent\textbf{Hyperparameters.} The training parameters are identical for every dataset. The student model is trained with SGD at a learning rate of 0.04 for training the source-only model and 0.0025 for target-domain training. The batch size is equal to 4.  The teacher's EMA update rate is set to $\alpha=0.9996$, and the confidence threshold at $\tau = 0.8$. The AP/mAP scores are reported at 0.5 IoU threshold, following previous works. The experiments were conducted on a single NVIDIA Tesla V100-SXM2 32 GiB GPU.

\noindent\textbf{Model selection.} Source-free student-teacher methods are subject to the issue of collapse. However, in a purely unsupervised scenario, the test set cannot be used for early stopping. We have observed that across the three tasks, the optimal early stopping for SF-UT is around 4000 training steps, hence, we report performance at this step. Most previous works \cite{chu_adversarial_2023,huang2021model,li2022source,li2021free,wei_entropy-minimization_2023,zhang2023refined} do not specify any model selection strategy and do not report learning curves, hinting towards the selection the best-performing iteration during training on the test set. While some works try to stabilize the training, these still require to set some hyper-parameters empirically. In real-world applications, a small labeled test set could be used for this purpose. Moreover, we show that our proposed AdaBN+Fixed SF-FM method has stable training, with similar performance.

%
\subsection{Importance of batch normalization}

\begin{table}
    \caption{Impact of batch normalization (BN) layers in the backbone (here, VGG16) on robustness to domain shift.}
    \label{tab:bn-ablation}
    \centering
    \begin{tabular}{lccccccc}
        \toprule
        \multirow{2}{*}{\textbf{Method}} & \multirow{2}{*}{\textbf{BN}} & \multicolumn{2}{c}{\textbf{City}$\rightarrow$\textbf{Foggy}} & \multicolumn{2}{c}{\textbf{KITTI}$\rightarrow$\textbf{City}} & \multicolumn{2}{c}{\textbf{SIM10k}$\rightarrow$\textbf{City}} \\
        & & source & target & source & target & source & target \\
        \midrule
        Source-only & \xmark & 46.3 & 20.0 & 86.6 & 14.6 & 40.2 & 31.0 \\
        Source-only & \cmark & 45.7 & 26.9 & 86.6 & 29.1 & 40.9 & 31.5 \\
        AdaBN & \cmark & - & 35.1 & - & 37.5 & - & 46.9 \\
        \bottomrule
    \end{tabular}
\end{table}

Table~\ref{tab:bn-ablation} exhibits the important role of BN layers in the backbone in each adaptation scenario. Although the source-trained models with or without BN have similar performance on the source domain, their performance differs drastically on the target domain. Firstly, the source-trained models with BN are more robust to domain shifts, with significantly higher performance on Cityscapes$\rightarrow$Foggy (26.9 vs 20.0 mAP with and without BN layers, respectively) and KITTI$\rightarrow$Cityscapes (29.1 vs 14.6 AP on car). Adapting the batch statistics on the target training data using AdaBN brings an additional significant performance gain. For instance, it reaches 35.1 mAP on Cityscapes$\rightarrow$Foggy, which is higher than many SFOD methods in literature. In the case of SIM10k$\rightarrow$Cityscapes where BN layers do not bring a significant improvement out-of-the-box (31.5 vs 31.0 AP on car), AdaBN still yields a large absolute improvement of +15.4 AP on car, outperforming most state-of-the-art SFOD methods.

\subsection{Regression losses}

\begin{table}
    \centering
    \caption{Ablation study on enabling the regression losses on unlabeled target samples during adaptation (Cityscapes$\rightarrow$Foggy).}
    \label{tab:rpn-reg}
    \begin{tabular}{lccccccccc|c}
        \toprule
        Method & $\mathcal{L}_{reg}$ & Person & Rider & Car & Truck & Bus & Train & Motor & Bicycle & mAP \\
        \midrule
        SF-UT & \xmark & 39.5 & 44.9 & 57.7 & \textbf{30.6} & 49.5 & 49.4 & 34.2 & 41.4 & 43.3 \\
        SF-UT & \cmark & \textbf{40.9} & \textbf{48.0} & \textbf{58.9} & 29.6 & \textbf{51.9} & \textbf{50.2} & \textbf{36.2} & \textbf{44.1} & \textbf{45.0} \\
        \bottomrule
    \end{tabular}
\end{table}

As mentioned previously, previous methods did not consider fine-tuning the detector regression losses, due to the lack of confidence estimates for the bounding box regression pseudo-labels. However, our ablation study shown Table~\ref{tab:rpn-reg} highlights a slightly higher performance when enabling the regression losses.

\subsection{Benchmark results}

\begin{table}[t]
    \centering
    \caption{Detection results on weather adaptation (Cityscapes$\rightarrow$Foggy-Cityscapes). Best and second-best results in bold.
    }
    \label{tab:citytofoggy}
    \resizebox{\linewidth}{!}{
    \begin{tabular}{cllcccccccc|c}
        \toprule
        Setting & Methods & Backbone & Person & Rider & Car & Truck & Bus & Train & Motor & Bicycle & mAP \\
        \midrule
        & Source-only & VGG16 & 20.3 & 27.2 & 30.3 & 12.7 & 17.6 & 4.9 & 28.3 & 27.3 & 20.0 \\
        & Source-only & VGG16-BN &  31.7 & 38.5 & 38.9 & 18.2 & 28.4 & 4.9 & 23.2 & 32.0 & 26.9 \\
        \midrule
        \multirow{9}*{UDAOD} & DA-Faster \cite{chen2018domain} & VGG16 & 25.0 & 31.0 & 40.5 & 22.1 & 35.3 & 20.2 & 20.0 & 27.1 & 27.6 \\
        & Pseudo-Label \cite{khodabandeh2019robust} & InceptionV2 & 31.9 & 39.9 & 48.0 & 25.1 & 39.9 & 27.2 & 25.0 & 34.1 & 33.9 \\
        & NL \cite{khodabandeh2019robust} & InceptionV2 & 35.1 & 42.2 & 49.2 & 30.1 & 45.3 & 27.0 & 26.9 & 36.0 & 36.5 \\
        & CDN \cite{su2020adapting} & VGG16 & 35.8 & 45.7 & 50.9 & 30.1 & 42.5 & 29.8 & 30.8 & 36.5 & 36.6 \\
        & ATF \cite{he2020domain} & VGG16 & 34.6 & 47.0 & 50.0 & 23.7 & 43.3 & 38.7 & 33.4 & 38.8 & 38.7 \\
        & MeGA-CDA \cite{vs2021mega} & VGG16 & 25.4 & 52.4 & 49.0 & 37.7 & 46.9 & 34.5 & 39.0 & 49.2 & 41.8 \\
        & UMT \cite{deng2021unbiased} & VGG16 & 33.0 & 46.7 & 48.6 & 34.1 & 56.5 & 46.8 & 30.4 & 37.3 & 41.7 \\
        & TIA \cite{zhao_task-specific_2022} & VGG16 & 31.1 & 49.7 & 46.3 & 34.8 & 48.6 & 37.7 & 38.1 & 52.1 & 42.3 \\
        & AT \cite{li2022cross} & VGG16-BN & 56.3 & 51.9 & 64.2 & 38.5 & 45.5 & 55.1 & 54.3 & 35.0 & 50.9 \\
        \midrule
        \multirow{13}*{SFOD}
        & SED \cite{li2021free} & VGG16 & 32.6 & 40.4 & 44.0 & 21.7 & 34.3 & 11.8 & 25.3 & 34.5 & 30.6 \\
        & SED-Mosaic \cite{li2021free} & VGG16 & 34.0 & 40.7 & 44.5 & 25.5 & 39.0 & 22.2 & 28.4 & 34.1 & 33.5 \\
        & HCL \cite{huang2021model} & VGG16 & 38.7 & 46.0 & 47.9 & \textbf{33.0} & 45.7 & 38.9 & 32.8 & 34.9 & 39.7 \\
        & SOAP \cite{xiong2021source} & VGG16 & 35.9 & 45.0 & 48.4 & 23.9 & 37.2 & 24.3 & 31.8 & 37.9 & 35.5 \\
        & LODS \cite{li2022source} & VGG16 & 34.0 & 45.7 & 48.8 & 27.3 & 39.7 & 19.6 & 33.2 & 37.8 & 35.8 \\
        & ESOD \cite{wei_entropy-minimization_2023} & ResNet101 & 28.3 & 39.3 & 49.3 & 28.8 & 31.5 & 21.7 & 24.7 & 31.5 & 31.9 \\
        & A$^2$SFOD \cite{chu_adversarial_2023} & VGG16 & 32.3 & 44.1 & 44.6 & 28.1 & 34.3 & 29.0 & 31.8 & 38.9 & 35.4 \\
        & IRG \cite{vs_instance_2023} & ResNet50 & 37.4 & 45.2 & 51.9 & 24.4 & 39.6 & 25.2 & 31.5 & 41.6 & 37.1 \\
        & RPL \cite{zhang2023refined} & VGG16 & 37.9 & \textbf{52.7} & 52.0 & 29.9 & 46.7 & 21.7 & 34.5 & \textbf{46.5} & \textbf{40.2} \\
        & Chen \textit{et al.} \cite{chen_exploiting_2023} & VGG16 & 39.0 & 50.3 & 55.4 & 24.0 & 46.0 & 21.2 & 30.3 & 44.2 & 38.8 \\
        & PETS \cite{liu_periodically_2023} (single level) & VGG16 & \textbf{42.0} & 48.7 & 56.3 & 19.3 & 39.3 & 5.5 & 34.2 & 41.6 & 35.9 \\
        \cmidrule{2-12}
        & AdaBN \cite{li2018adaptive} & VGG16-BN & 38.5 & 44.1 & 49.9 & 26.0 & 39.2 & 16.6 & 29.0 & 37.4 & 35.1 \\
        & SF-UT (Ours) & VGG16 & 36.5 & 44.6 & 54.8 & 30.2 & 45.3 & 31.8 & 29.5 & 41.0 & 39.2 \\
        \rowcolor[gray]{0.85}&SF-UT (Ours) & VGG16-BN & 40.9 & 48.0 & \textbf{58.9} & 29.6 & \textbf{51.9} & \textbf{50.2} & \textbf{36.2} & 44.1 & \textbf{45.0} \\
        \midrule
        & Oracle & VGG16 & 51.3 & 57.5 & 70.2 & 30.9 & 60.5 & 26.9 & 40.0 & 50.4 & 43.5 \\
        \bottomrule
    \end{tabular}
    }
\end{table}

\begin{table}
    \centering
    \begin{minipage}[t]{0.49\linewidth}
    \vspace{0pt}
    \caption{Detection results on cross-camera adaptation (KITTI$\rightarrow$Cityscapes). Best and second-best results in bold.}
    \label{tab:kittitocity}
    \resizebox{\linewidth}{!}{
    \begin{tabular}{cllc}
        \toprule
        Setting & Methods & Backbone & AP on car \\
        \midrule
        & Source-only & VGG16 & 14.6\\
        & Source-only & VGG16-BN & 29.1 \\
        \midrule
        \multirow{9}*{UDAOD} & DA-Faster \cite{chen2018domain} & VGG16 & 38.5 \\
        & SW-Faster \cite{saito_strong-weak_2019} & VGG16 & 37.9 \\
        & Pseudo-Label \cite{khodabandeh2019robust} & InceptionV2 & 40.2 \\
        & NL \cite{khodabandeh2019robust} & InceptionV2 & 43.0 \\
        & ATF \cite{he2020domain} & VGG16 & 42.1 \\
        & SAPNet \cite{li2020spatial} & VGG16 & 43.4 \\
        & CST-DA \cite{zhao2020collaborative} & VGG16 & 43.6 \\
        & MeGA-CDA \cite{vs2021mega} & VGG16 & 43.0 \\
        & SGA-S \cite{zhang2021self} & ResNet101& 43.5 \\
        \midrule
        \multirow{10}*{SFOD} & SED \cite{li2021free} & VGG16 & 43.6 \\
        & SED-Mosaic \cite{li2021free} & VGG16 & 44.6 \\
        & SOAP \cite{xiong2021source} & VGG16 & 42.7 \\
        & LODS \cite{li2022source} & VGG16 & 43.9 \\
        & A$^2$SFOD \cite{chu_adversarial_2023} & VGG16 & 44.9 \\
        & IRG \cite{vs_instance_2023} & ResNet50 & 45.7 \\
        & RPL \cite{zhang2023refined} & VGG16 & \textbf{47.8} \\
        & PETS \cite{liu_periodically_2023} & VGG16 & \textbf{47.0} \\
        \cmidrule{2-4}
        & AdaBN \cite{li2018adaptive} & VGG16-BN & 37.5 \\
        \rowcolor[gray]{0.85}&SF-UT (Ours) & VGG16-BN & 46.2 \\
        \midrule
        & Oracle & VGG16 & 49.9 \\
        \bottomrule
    \end{tabular}
    }
    \end{minipage}
    \begin{minipage}[t]{0.49\linewidth}
    \vspace{0pt}
    \caption{Detection results on synthetic-to-real adaptation (SIM10k$\rightarrow$Cityscapes). Best and second-best results in bold.}
    \label{tab:sim10ktocity}
    \resizebox{\linewidth}{!}{
    \begin{tabular}{cllc}
        \toprule
        Setting & Methods & Backbone & AP on car \\
        \midrule
        & Source-only & VGG16 & 31.0 \\
        & Source-only & VGG16-BN & 31.5 \\
        \midrule
        \multirow{6}*{UDAOD} & DA-Faster \cite{chen2018domain} & VGG16 & 38.9 \\
        & Pseudo-Label \cite{khodabandeh2019robust} & InceptionV2 & 39.1 \\ 
        & NL \cite{khodabandeh2019robust} & InceptionV2 & 42.6 \\
        & CDN \cite{su2020adapting} & VGG16 & 49.3 \\
        & ATF \cite{he2020domain} & VGG16 & 42.8 \\
        & UMT \cite{deng2021unbiased} & VGG16 & 43.1 \\
        \midrule
        \multirow{11}*{SFOD} & SED \cite{li2021free} & VGG16 & 42.3 \\
        & SED-Mosaic \cite{li2021free} & VGG16 & 42.9 \\
        & SOAP \cite{xiong2021source} & VGG16 & 40.8 \\
        & LODS \cite{li2022source} & VGG16 & 45.7 \\
        & ESOD \cite{wei_entropy-minimization_2023} & ResNet101 & 42.4 \\
        & A$^2$SFOD \cite{chu_adversarial_2023} & VGG16 & 44.0 \\
        & IRG \cite{vs_instance_2023} & ResNet50 & 43.2 \\
        & RPL \cite{zhang2023refined} & VGG16 & 50.1 \\
        & PETS \cite{liu_periodically_2023} & VGG16 & \textbf{57.8} \\
        \cmidrule{2-4}
        & AdaBN \cite{li2018adaptive} & VGG16-BN &  46.9 \\
        \rowcolor[gray]{0.85}&SF-UT (Ours) & VGG16-BN & \textbf{55.4} \\
        \midrule
        & Oracle & VGG16 & 58.5 \\ 
        \bottomrule
    \end{tabular}
    }
    \end{minipage}
\end{table}

In this section, we compare our proposed SF-UT with various state-of-the-art methods for UDAOD and SFOD. Results are displayed in Table~\ref{tab:citytofoggy}, Table~\ref{tab:kittitocity} and Table~\ref{tab:sim10ktocity} for the three adaptation tasks. We also indicate the backbone in each method, according to the paper and available code. Note that for VGG16 backbones, the presence of BN is not always indicated by the authors.

Among all compared SFOD methods, SF-UT exhibits the best performance on Cityscapes$\rightarrow$Foggy with a mAP of 45.0\%, outperforming the previous state-of-the-art RPL \cite{zhang2023refined} by a large margin. This is partly due to the superiority of the VGG16-BN backbone. Thus, we also experimented with the VGG16 backbone used in previous methods. In this case, SF-UT reaches 39.2 mAP, which is still close to state-of-the-art. However, SF-UT is much simpler and comprises none of the regularization, alignment or pseudo-label enhancement techniques used in other more complex methods. On KITTI$\rightarrow$Cityscapes, SF-UT is also close to the best-performing methods PETS \cite{liu_periodically_2023} and RPL \cite{zhang2023refined}. Finally, on the SIM10k$\rightarrow$Cityscapes, SF-UT is only surpassed by PETS and outperforms other methods by a large margin.



%
\subsection{Exploration of self-training strategies}

\begin{table}
    \centering
    \caption{Study on various self-training configurations for source-free object detection. \textit{Fixed PLs} indicates the set of target pseudo-labels is fixed during training. \textit{Weak-Strong} means weakly augmented inputs provide supervision for strongly-augmented inputs. \textit{AdaBN} means that the batch statistics were adapted on the target training data before generating a set of fixed PLs. Results on weather adaptation.}
    \label{tab:exploration}
    \resizebox{\textwidth}{!}{
    \begin{tabular}{lcccccccccc|c}
        \toprule
       Methods & Fixed PLs & Weak-Strong & Person & Rider & Car & Truck & Bus & Train & Motor & Bicycle & mAP \\
        \midrule
        AdaBN & - & - & 38.5 & 44.1 & 49.9 & 26.0 & 39.2 & 16.6 & 29.0 & 37.4 & 35.1 \\
        SF-Pseudo-Label & \xmark & \xmark & 39.7 & 45.6 & 54.0 & 26.0 & 41.2 & 32.2 & 30.3 & 40.3 & 38.7 \\
        SF-FixMatch & \xmark & \cmark & 37.7 & 45.2 & 54.2 & 25.2 & 42.0 & 20.7 & 30.0 & 40.0 & 36.8 \\
        \midrule
        Fixed SF-Pseudo-Label  & \cmark & \xmark & 38.5 & 42.8 & 48.1 & 25.2 & 37.3 & 26.5 & 28.8 & 37.6 & 35.6\\
        Fixed SF-FixMatch  & \cmark & \cmark & 38.7 & 45.8 & 50.9 & 29.1 &  36.9 & 38.4 & 33.1 &  39.1 & 39.0 \\
        AdaBN + Fixed SF-PL  & \cmark & \xmark & 40.0 & 45.1 & 52.3 & 26.9 & 43.7 & 28.2 & 31.0 & 41.2 & 38.5 \\
        \rowcolor[gray]{0.85}AdaBN + Fixed SF-FM  & \cmark & \cmark & \textbf{41.5} & \textbf{48.1} & 55.1 & \textbf{30.8} & 49.3 & \textbf{53.0} & 34.2 & \textbf{44.1} & 44.5\\
        \midrule
        Mean Teacher & \xmark & \xmark & 39.0 & 46.3 & 53.8 & 26.6 & 43.2 & 27.7 & 28.3 & 40.5 & 38.2 \\
        \rowcolor[gray]{0.85}SF-UT  & \xmark & \cmark & 40.9 & 48.0 & \textbf{58.9} & 29.6 & \textbf{51.9} & 50.2 & \textbf{36.2} & \textbf{44.1}& \textbf{45.0} \\
        \midrule
        \midrule
        AdaBN + Fixed SF-PL + Mosaic  & \cmark & \xmark & 40.7 & 47.2 & 55.5 & 28.1 & 43.7 & 37.1 & 31.0 & 42.0 & 40.7 \\
        \rowcolor[gray]{0.85}AdaBN + Fixed SF-FM + Mosaic & \cmark & \cmark & \textbf{41.9} & \textbf{48.6} & 57.3 & \textbf{33.0} & 46.7 & \textbf{53.4} & 33.8 & 42.2 & \textbf{44.6} \\
        \bottomrule
    \end{tabular}
    }
\end{table}

\begin{figure}
    \centering
    \includegraphics[width=0.99\linewidth]{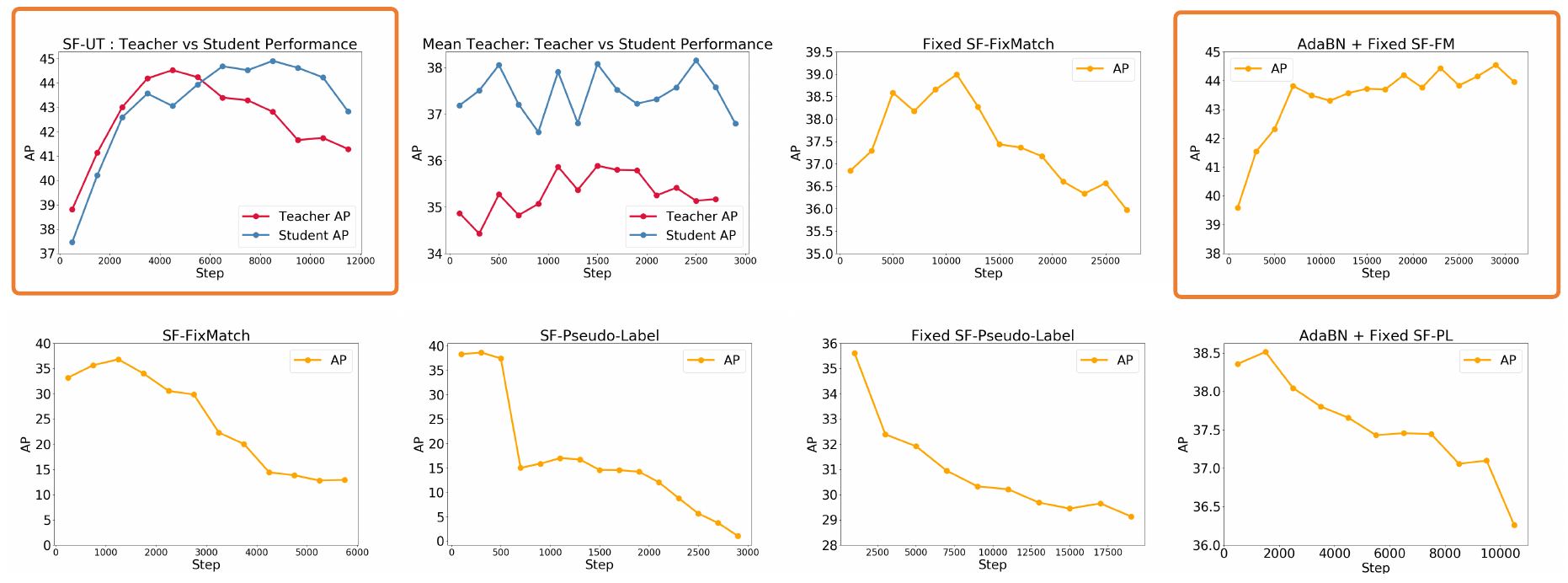}
    \caption{Training curves of the studied self-training strategies on Cityscapes$\rightarrow$Foggy. The best-performing ones are the proposed Source-Free Unbiased Teacher (SF-UT) and AdaBN + Fixed Source-Free FixMatch (SF-FM), the latter being the only method not subject to collapse during training.}
    \label{fig:curves}
\end{figure}

In this section, we explore the various configurations of self-training described in Section~\ref{sec:method} and evaluate the performance of the source-free variants of Pseudo-Label and FixMatch. Results are reported in Table~\ref{tab:exploration} for Cityscapes$\rightarrow$Foggy-Cityscapes. The two main observations are the importance of weak-strong augmentation, and the ability of fixed pseudo-labels to reach higher performance by preventing collapse. The Fixed SF-PL and Fixed SF-FM methods achieve competitive results with respectively 35.6 and 39.0 mAP. Without fixed PLs, the performance quickly collapses (see Figure~\ref{fig:curves}).
Remarkably, AdaBN + Fixed SF-FM reaches similar performance to SF-UT, with a mAP of 44.5. The higher quality of the fixed set of initial pseudo-labels via AdaBN lead to a +5.5 absolute improvement in mAP. For Fixed SF-PL, AdaBN yielded an improvement of +2.9, reaching 38.5 mAP. These findings challenge the superiority of mutual teacher-student learning, the dominant approach for SFOD in this scenario. Moreover, AdaBN + Fixed SF-FM is the only approach with a stable training, as can be seen on its learning curve on Figure~\ref{fig:curves}. In other words, to obtain high-quality PLs and prevent model collapse, using fixed PLs is only effective when used in combination with AdaBN (to mitigate domain shift) and weak-strong augmentations, while the various ablations of this method still suffer from model collapse.

Inspired by the large improvement due to strong augmentation on the inputs, we try to further augment the target input images with Mosaic augmentation \cite{yolox2021} applied on top of the strong augmentation. Four strongly-augmented images from the target domain are concatenated together to form a mosaic and then resized into a single input image. The performance of AdaBN + Fixed SF-PL and AdaBN + Fixed SF-FM with Mosaic augmentation is further improved to 40.7 and 44.6 mAP respectively.

\begin{table}[h]
    \begin{minipage}[t]{0.49\linewidth}
    \vspace{0pt}
    \setlength{\aboverulesep}{0pt}
    \setlength{\belowrulesep}{0pt}
    \centering
    \caption{AdaBN + Fixed SF-FM detection results on cross-camera adaptation (KITTI$\rightarrow$Cityscapes). }
    \label{tab:kittitocity2}
    \resizebox{\linewidth}{!}{
    \begin{tabular}{llc}
        \toprule
        Methods & Backbone & AP on car \\
        \midrule
        Source-only & VGG16-BN & 29.1 \\
        \midrule
        AdaBN & VGG16-BN & 37.5 \\
        \rowcolor[gray]{0.85}AdaBN + Fixed SF-FM (Ours) & VGG16-BN & 45.2 \\
        \rowcolor[gray]{0.85}SF-UT (Ours) & VGG16-BN & \textbf{46.2} \\
        \midrule
        Oracle & & 49.9 \\
        \bottomrule
    \end{tabular}
    }
    \end{minipage}
    \begin{minipage}[t]{0.49\linewidth}
    \vspace{0pt}
    \setlength{\aboverulesep}{0pt}
    \setlength{\belowrulesep}{0pt}
    \centering
    \caption{AdaBN + Fixed SF-FM detection results on synthetic-to-real adaptation (SIM10k$\rightarrow$Cityscapes).}
    \label{tab:sim10ktocity2}
    \resizebox{\linewidth}{!}{
    \begin{tabular}{llc}
        \toprule
        Methods & Backbone & AP on car \\
        \midrule
        Source-only & VGG16-BN & 31.5\\
        \midrule
        AdaBN & VGG16-BN & 46.9 \\
        \rowcolor[gray]{0.85}AdaBN + Fixed SF-FM (Ours) & VGG16-BN & 53.3 \\
        \rowcolor[gray]{0.85}SF-UT (Ours) & VGG16-BN & \textbf{55.4}\\
        \midrule
        Oracle & & 58.5 \\ 
        \bottomrule
    \end{tabular}
    }
\end{minipage}
\end{table}

Furthermore, Table~\ref{tab:kittitocity2} and \ref{tab:sim10ktocity2} also report results on KITTI$\rightarrow$Cityscapes and SIM10k$\rightarrow$Cityscapes respectively, for the strategy AdaBN + Fixed SF-FM training on fixed pseudo-labels with weak-strong augmentations, after AdaBN adaptation. As on Foggy-Cityscapes, this strategy is effective, with performance very close to SF-UT (Source-Free Unbiased Teacher), while being simpler.

\section{Conclusion}

In this study, we have investigated and evaluated simple yet effective approaches to source-free domain adaptation for object detection. After showing the importance of batch normalization and the effectiveness of AdaBN, we proposed a Source-Free Unbiased Teacher (SF-UT), achieving state-of-the-art performance on Foggy-Cityscapes and competitive results on other benchmarks. Moreover, we introduced a simple strategy consisting in training on a fixed set of pseudo-labels with strong augmentations after batch statistics adaptation (AdaBN + Fixed SF-FM), also yielding promising performance and notably mitigating the issue of collapse in self-training. Overall, we have shown how simpler approaches could actually exceed previous, much more complex, SFOD methods. 

A limitation of our work, which concerns all domain adaptation methods relying on Batch Normalization, is its restriction to models incorporating BatchNorm layers. However, many modern architectures such as ConvNeXt or Transformers, favor GroupNorm or LayerNorm. The adaptation of GN and LN layers for domain adaptation is, to our knowledge, unexplored, and is an interesting research perspective. 
As part of future work, we also consider evaluating various different backbones and detectors.




\section*{Acknowledgements}

This work has been supported by the SNSF Grant 200021\_200461.

\bibliographystyle{splncs04}
\bibliography{main}

\begin{thebibliography}{10}
\providecommand{\url}[1]{\texttt{#1}}
\providecommand{\urlprefix}{URL }
\providecommand{\doi}[1]{https://doi.org/#1}

\bibitem{cai2019exploring}
Cai, Q., Pan, Y., Ngo, C.W., Tian, X., Duan, L., Yao, T.: Exploring object relation in mean teacher for cross-domain detection. In: Proceedings of the IEEE/CVF Conference on Computer Vision and Pattern Recognition. pp. 11457--11466 (2019)

\bibitem{chang_domain-specific_2019}
Chang, W.G., You, T., Seo, S., Kwak, S., Han, B.: Domain-{Specific} {Batch} {Normalization} for {Unsupervised} {Domain} {Adaptation}. In: 2019 {IEEE}/{CVF} {Conference} on {Computer} {Vision} and {Pattern} {Recognition} ({CVPR}). pp. 7346--7354 (Jun 2019). \doi{10.1109/CVPR.2019.00753}, \url{https://ieeexplore.ieee.org/document/8953938}, iSSN: 2575-7075

\bibitem{chen2018domain}
Chen, Y., Li, W., Sakaridis, C., Dai, D., Van~Gool, L.: Domain adaptive faster r-cnn for object detection in the wild. In: Proceedings of the IEEE conference on computer vision and pattern recognition. pp. 3339--3348 (2018)

\bibitem{chen_exploiting_2023}
Chen, Z., Wang, Z., Zhang, Y.: Exploiting {Low}-confidence {Pseudo}-labels for {Source}-free {Object} {Detection} (Oct 2023). \doi{10.1145/3581783.3612273}, \url{http://arxiv.org/abs/2310.12705}, arXiv:2310.12705 [cs]

\bibitem{chu_adversarial_2023}
Chu, Q., Li, S., Chen, G., Li, K., Li, X.: Adversarial {Alignment} for {Source} {Free} {Object} {Detection} (Jan 2023). \doi{10.48550/arXiv.2301.04265}, \url{http://arxiv.org/abs/2301.04265}, arXiv:2301.04265 [cs]

\bibitem{Cordts2016Cityscapes}
Cordts, M., Omran, M., Ramos, S., Rehfeld, T., Enzweiler, M., Benenson, R., Franke, U., Roth, S., Schiele, B.: The cityscapes dataset for semantic urban scene understanding. In: Proc. of the IEEE Conference on Computer Vision and Pattern Recognition (CVPR) (2016)

\bibitem{courty_joint_2017}
Courty, N., Flamary, R., Habrard, A., Rakotomamonjy, A.: Joint distribution optimal transportation for domain adaptation. In: Advances in {Neural} {Information} {Processing} {Systems}. vol.~30. Curran Associates, Inc. (2017), \url{https://proceedings.neurips.cc/paper/2017/hash/0070d23b06b1486a538c0eaa45dd167a-Abstract.html}

\bibitem{deng2021unbiased}
Deng, J., Li, W., Chen, Y., Duan, L.: Unbiased mean teacher for cross-domain object detection. In: Proceedings of the IEEE/CVF Conference on Computer Vision and Pattern Recognition. pp. 4091--4101 (2021)

\bibitem{duan2012domain}
Duan, L., Tsang, I.W., Xu, D.: Domain transfer multiple kernel learning. IEEE Transactions on Pattern Analysis and Machine Intelligence  \textbf{34}(3),  465--479 (2012)

\bibitem{ganin_domain-adversarial_2016}
Ganin, Y., Ustinova, E., Ajakan, H., Germain, P., Larochelle, H., Laviolette, F., March, M., Lempitsky, V.: Domain-{Adversarial} {Training} of {Neural} {Networks}. Journal of Machine Learning Research  \textbf{17}(59),  1--35 (2016), \url{http://jmlr.org/papers/v17/15-239.html}

\bibitem{yolox2021}
Ge, Z., Liu, S., Wang, F., Li, Z., Sun, J.: Yolox: Exceeding yolo series in 2021. arXiv preprint arXiv:2107.08430  (2021)

\bibitem{Geiger2012CVPR}
Geiger, A., Lenz, P., Urtasun, R.: Are we ready for autonomous driving? the kitti vision benchmark suite. In: Conference on Computer Vision and Pattern Recognition (CVPR) (2012)

\bibitem{girshick2015fast}
Girshick, R.: Fast r-cnn. In: Proceedings of the IEEE international conference on computer vision. pp. 1440--1448 (2015)

\bibitem{gong2012geodesic}
Gong, B., Shi, Y., Sha, F., Grauman, K.: Geodesic flow kernel for unsupervised domain adaptation. In: 2012 IEEE conference on computer vision and pattern recognition. pp. 2066--2073. IEEE (2012)

\bibitem{he2020domain}
He, Z., Zhang, L.: Domain adaptive object detection via asymmetric tri-way faster-rcnn. In: Computer Vision--ECCV 2020: 16th European Conference, Glasgow, UK, August 23--28, 2020, Proceedings, Part XXIV 16. pp. 309--324. Springer (2020)

\bibitem{huang2021model}
Huang, J., Guan, D., Xiao, A., Lu, S.: Model adaptation: Historical contrastive learning for unsupervised domain adaptation without source data. Advances in Neural Information Processing Systems  \textbf{34},  3635--3649 (2021)

\bibitem{ioffe_batch_2015}
Ioffe, S., Szegedy, C.: Batch {Normalization}: {Accelerating} {Deep} {Network} {Training} by {Reducing} {Internal} {Covariate} {Shift} (Mar 2015). \doi{10.48550/arXiv.1502.03167}, \url{http://arxiv.org/abs/1502.03167}, arXiv:1502.03167 [cs]

\bibitem{johnson2017driving}
Johnson-Roberson, M., Barto, C., Mehta, R., Sridhar, S.N., Rosaen, K., Vasudevan, R.: Driving in the matrix: Can virtual worlds replace human-generated annotations for real world tasks? In: 2017 IEEE International Conference on Robotics and Automation (ICRA). pp. 746--753. IEEE (2017)

\bibitem{khodabandeh2019robust}
Khodabandeh, M., Vahdat, A., Ranjbar, M., Macready, W.G.: A robust learning approach to domain adaptive object detection. In: Proceedings of the IEEE/CVF International Conference on Computer Vision. pp. 480--490 (2019)

\bibitem{klingner_unsupervised_2021}
Klingner, M., Termöhlen, J.A., Ritterbach, J., Fingscheidt, T.: Unsupervised {BatchNorm} {Adaptation} ({UBNA}): {A} {Domain} {Adaptation} {Method} for {Semantic} {Segmentation} {Without} {Using} {Source} {Domain} {Representations} (Nov 2021). \doi{10.48550/arXiv.2011.08502}, \url{http://arxiv.org/abs/2011.08502}, arXiv:2011.08502 [cs]

\bibitem{laine_temporal_2017}
Laine, S., Aila, T.: Temporal {Ensembling} for {Semi}-{Supervised} {Learning} (Mar 2017). \doi{10.48550/arXiv.1610.02242}, \url{http://arxiv.org/abs/1610.02242}, arXiv:1610.02242 [cs]

\bibitem{lee_pseudo-label_2013}
Lee, D.H.: Pseudo-{Label} : {The} {Simple} and {Efficient} {Semi}-{Supervised} {Learning} {Method} for {Deep} {Neural} {Networks}. ICML 2013 Workshop : Challenges in Representation Learning (WREPL)  (Jul 2013)

\bibitem{li2020spatial}
Li, C., Du, D., Zhang, L., Wen, L., Luo, T., Wu, Y., Zhu, P.: Spatial attention pyramid network for unsupervised domain adaptation. In: Computer Vision--ECCV 2020: 16th European Conference, Glasgow, UK, August 23--28, 2020, Proceedings, Part XIII 16. pp. 481--497. Springer (2020)

\bibitem{li2022source}
Li, S., Ye, M., Zhu, X., Zhou, L., Xiong, L.: Source-free object detection by learning to overlook domain style. In: Proceedings of the IEEE/CVF Conference on Computer Vision and Pattern Recognition. pp. 8014--8023 (2022)

\bibitem{li2021free}
Li, X., Chen, W., Xie, D., Yang, S., Yuan, P., Pu, S., Zhuang, Y.: A free lunch for unsupervised domain adaptive object detection without source data. In: Proceedings of the AAAI Conference on Artificial Intelligence. vol.~35, pp. 8474--8481 (2021)

\bibitem{li2018adaptive}
Li, Y., Wang, N., Shi, J., Hou, X., Liu, J.: Adaptive batch normalization for practical domain adaptation. Pattern Recognition  \textbf{80},  109--117 (2018)

\bibitem{li2022cross}
Li, Y.J., Dai, X., Ma, C.Y., Liu, Y.C., Chen, K., Wu, B., He, Z., Kitani, K., Vajda, P.: Cross-domain adaptive teacher for object detection. In: Proceedings of the IEEE/CVF Conference on Computer Vision and Pattern Recognition. pp. 7581--7590 (2022)

\bibitem{liu_periodically_2023}
Liu, Q., Lin, L., Shen, Z., Yang, Z.: Periodically {Exchange} {Teacher}-{Student} for {Source}-{Free} {Object} {Detection}. In: {ICCV} 2023 (2023)

\bibitem{liu2021unbiased}
Liu, Y.C., Ma, C.Y., He, Z., Kuo, C.W., Chen, K., Zhang, P., Wu, B., Kira, Z., Vajda, P.: Unbiased teacher for semi-supervised object detection. arXiv preprint arXiv:2102.09480  (2021)

\bibitem{long2015learning}
Long, M., Cao, Y., Wang, J., Jordan, M.: Learning transferable features with deep adaptation networks. In: International conference on machine learning. pp. 97--105. PMLR (2015)

\bibitem{long_conditional_2018}
Long, M., CAO, Z., Wang, J., Jordan, M.I.: Conditional {Adversarial} {Domain} {Adaptation}. In: Advances in {Neural} {Information} {Processing} {Systems}. vol.~31. Curran Associates, Inc. (2018), \url{https://proceedings.neurips.cc/paper/2018/hash/ab88b15733f543179858600245108dd8-Abstract.html}

\bibitem{pan_two_2018}
Pan, X., Luo, P., Shi, J., Tang, X.: Two at {Once}: {Enhancing} {Learning} and {Generalization} {Capacities} via {IBN}-{Net} (2018), \url{https://openaccess.thecvf.com/content_ECCV_2018/html/Xingang_Pan_Two_at_Once_ECCV_2018_paper.html}

\bibitem{rangwani_closer_2022}
Rangwani, H., Aithal, S.K., Mishra, M., Jain, A., Babu, R.V.: A {Closer} {Look} at {Smoothness} in {Domain} {Adversarial} {Training} (Jun 2022). \doi{10.48550/arXiv.2206.08213}, \url{http://arxiv.org/abs/2206.08213}, arXiv:2206.08213 [cs]

\bibitem{ren2015faster}
Ren, S., He, K., Girshick, R., Sun, J.: Faster r-cnn: Towards real-time object detection with region proposal networks. Advances in neural information processing systems  \textbf{28} (2015)

\bibitem{saito_strong-weak_2019}
Saito, K., Ushiku, Y., Harada, T., Saenko, K.: Strong-{Weak} {Distribution} {Alignment} for {Adaptive} {Object} {Detection}. In: 2019 {IEEE}/{CVF} {Conference} on {Computer} {Vision} and {Pattern} {Recognition} ({CVPR}). pp. 6949--6958. IEEE, Long Beach, CA, USA (2019). \doi{10.1109/CVPR.2019.00712}, \url{https://ieeexplore.ieee.org/document/8954336/}

\bibitem{sakaridis_semantic_2018}
Sakaridis, C., Dai, D., Van~Gool, L.: Semantic {Foggy} {Scene} {Understanding} with {Synthetic} {Data}. International Journal of Computer Vision  \textbf{126}(9),  973--992 (Sep 2018). \doi{10.1007/s11263-018-1072-8}, \url{http://link.springer.com/10.1007/s11263-018-1072-8}

\bibitem{seo_learning_2020}
Seo, S., Suh, Y., Kim, D., Kim, G., Han, J., Han, B.: Learning to {Optimize} {Domain} {Specific} {Normalization} for {Domain} {Generalization}. In: Vedaldi, A., Bischof, H., Brox, T., Frahm, J.M. (eds.) Computer {Vision} – {ECCV} 2020. Lecture {Notes} in {Computer} {Science}, Springer International Publishing, Cham (2020). \doi{10.1007/978-3-030-58542-6_5}

\bibitem{simonyan2014very}
Simonyan, K., Zisserman, A.: Very deep convolutional networks for large-scale image recognition. arXiv preprint arXiv:1409.1556  (2014)

\bibitem{sohn2020fixmatch}
Sohn, K., Berthelot, D., Carlini, N., Zhang, Z., Zhang, H., Raffel, C.A., Cubuk, E.D., Kurakin, A., Li, C.L.: Fixmatch: Simplifying semi-supervised learning with consistency and confidence. Advances in neural information processing systems  \textbf{33},  596--608 (2020)

\bibitem{su2020adapting}
Su, P., Wang, K., Zeng, X., Tang, S., Chen, D., Qiu, D., Wang, X.: Adapting object detectors with conditional domain normalization. In: Computer Vision--ECCV 2020: 16th European Conference, Glasgow, UK, August 23--28, 2020, Proceedings, Part XI 16. pp. 403--419. Springer (2020)

\bibitem{tarvainen2017mean}
Tarvainen, A., Valpola, H.: Mean teachers are better role models: Weight-averaged consistency targets improve semi-supervised deep learning results. Advances in neural information processing systems  \textbf{30} (2017)

\bibitem{ulyanov_instance_2017}
Ulyanov, D., Vedaldi, A., Lempitsky, V.: Instance {Normalization}: {The} {Missing} {Ingredient} for {Fast} {Stylization} (Nov 2017). \doi{10.48550/arXiv.1607.08022}, \url{http://arxiv.org/abs/1607.08022}, arXiv:1607.08022 [cs]

\bibitem{vs2021mega}
Vs, V., Gupta, V., Oza, P., Sindagi, V.A., Patel, V.M.: Mega-cda: Memory guided attention for category-aware unsupervised domain adaptive object detection. In: Proceedings of the IEEE/CVF Conference on Computer Vision and Pattern Recognition. pp. 4516--4526 (2021)

\bibitem{vs_instance_2023}
VS, V., Oza, P., Patel, V.M.: Instance {Relation} {Graph} {Guided} {Source}-{Free} {Domain} {Adaptive} {Object} {Detection} (Mar 2023). \doi{10.48550/arXiv.2203.15793}, \url{http://arxiv.org/abs/2203.15793}, arXiv:2203.15793 [cs]

\bibitem{wang_tent_2021}
Wang, D., Shelhamer, E., Liu, S., Olshausen, B., Darrell, T.: Tent: {Fully} {Test}-time {Adaptation} by {Entropy} {Minimization} (Mar 2021). \doi{10.48550/arXiv.2006.10726}, \url{http://arxiv.org/abs/2006.10726}, arXiv:2006.10726 [cs, stat]

\bibitem{wei_entropy-minimization_2023}
Wei, X., Bai, T., Duan, Z., Zhao, M., Zhao, C., Lu, Y., Hu, D.: Entropy-minimization {Mean} {Teacher} for {Source}-{Free} {Domain} {Adaptive} {Object} {Detection}. In: Neural {Information} {Processing}, vol. 13623, pp. 513--524. Springer International Publishing, Cham (2023). \doi{10.1007/978-3-031-30105-6_43}, \url{https://link.springer.com/10.1007/978-3-031-30105-6_43}, series Title: Lecture Notes in Computer Science

\bibitem{xiong2021source}
Xiong, L., Ye, M., Zhang, D., Gan, Y., Li, X., Zhu, Y.: Source data-free domain adaptation of object detector through domain-specific perturbation. International Journal of Intelligent Systems  \textbf{36}(8),  3746--3766 (2021)

\bibitem{xu2020exploring}
Xu, C.D., Zhao, X.R., Jin, X., Wei, X.S.: Exploring categorical regularization for domain adaptive object detection. In: Proceedings of the IEEE/CVF Conference on Computer Vision and Pattern Recognition. pp. 11724--11733 (2020)

\bibitem{zhang2021self}
Zhang, C., Li, Z., Liu, J., Peng, P., Ye, Q., Lu, S., Huang, T., Tian, Y.: Self-guided adaptation: Progressive representation alignment for domain adaptive object detection. IEEE Transactions on Multimedia  \textbf{24},  2246--2258 (2021)

\bibitem{zhang_generalizable_2022}
Zhang, J., Qi, L., Shi, Y., Gao, Y.: Generalizable model-agnostic semantic segmentation via target-specific normalization. Pattern Recognition  \textbf{122} (2022). \doi{10.1016/j.patcog.2021.108292}, \url{https://www.sciencedirect.com/science/article/pii/S0031320321004726}

\bibitem{zhang2023refined}
Zhang, S., Zhang, L., Liu, Z.: Refined pseudo labeling for source-free domain adaptive object detection. arXiv preprint arXiv:2303.03728  (2023)

\bibitem{zhao2020collaborative}
Zhao, G., Li, G., Xu, R., Lin, L.: Collaborative training between region proposal localization and classification for domain adaptive object detection. In: Computer Vision--ECCV 2020: 16th European Conference, Glasgow, UK, August 23--28, 2020, Proceedings, Part XVIII 16. pp. 86--102. Springer (2020)

\bibitem{zhao_task-specific_2022}
Zhao, L., Wang, L.: Task-specific {Inconsistency} {Alignment} for {Domain} {Adaptive} {Object} {Detection} (Mar 2022). \doi{10.48550/arXiv.2203.15345}, \url{http://arxiv.org/abs/2203.15345}, arXiv:2203.15345 [cs]

\bibitem{zhao2020bi}
Zhao, Z., Guo, Y., Ye, J.: Bi-dimensional feature alignment for cross-domain object detection. In: Computer Vision--ECCV 2020 Workshops: Glasgow, UK, August 23--28, 2020, Proceedings, Part I 16. pp. 671--686. Springer (2020)

\end{thebibliography}
\clearpage
 \section*{Supplementary Material}

\section*{Example visualizations}

\begin{figure*}[!htbp]
    \centering
    \includegraphics[width=0.99\linewidth]{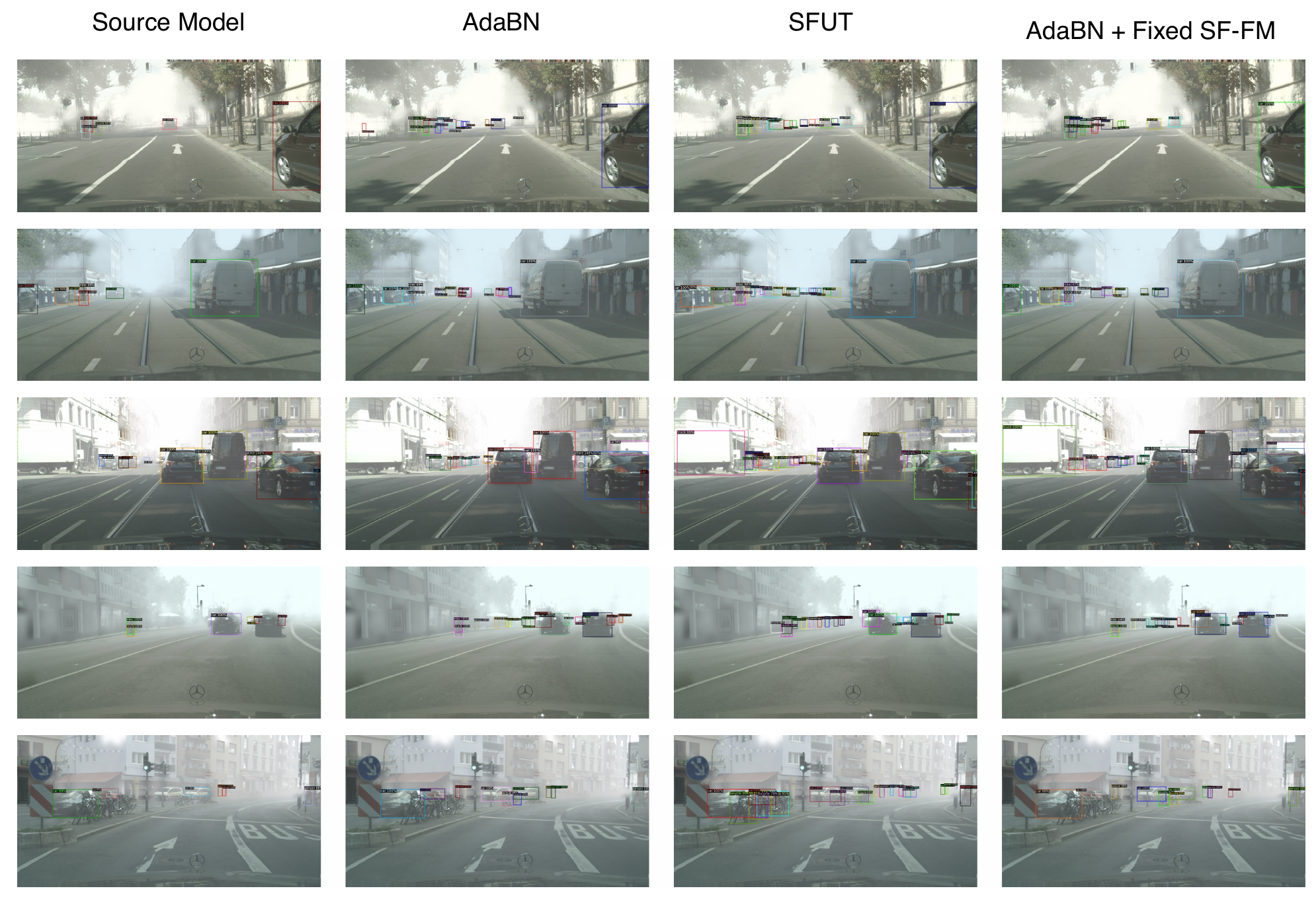}
    \caption{Illustration of the visualization results from different models. From left to right, they are source model, AdaBN model, SF-UT model and AdaBN + Fixed SF-FM. The adaptation task is from Cityscapes to Foggy-Cityscapes. The images shown are on Foggy-Cityscapes.}
    \label{fig:visual}
\end{figure*}

As the visualization results Figure \ref{fig:visual} (Cityscapes $\rightarrow$ Foggy-Cityscapes) illustrate, SFUT and AdaBN + Fixed SF-FM can efficiently detect objects that the source pre-trained model ignores, especially shrouded people in fog, stacked bicycles, and incomplete cars. AdaBN also shows a significantly better performance than the source model.

\section*{Code}

The code of this submission can be found at \href{https://github.com/EPFL-IMOS/simple-SFOD}{github.com/EPFL-IMOS/simple-SFOD}.

\end{document}